\documentclass[letterpaper, 10 pt, conference]{ieeeconf}  

\IEEEoverridecommandlockouts                              

\usepackage{amsmath}
\usepackage{amssymb}
\usepackage{mathrsfs}
\usepackage{mathdots}
\usepackage{bbm}
\usepackage{mathtools}
\usepackage{bm}
\usepackage{amsfonts}
\usepackage{cases}
\usepackage{graphicx}
\usepackage{cite}
\usepackage{multirow}
\usepackage{color}
\usepackage{algorithm}
\usepackage{algpseudocode}
\usepackage{graphicx}
\usepackage{subfigure}
\usepackage{epstopdf}
\usepackage{lipsum}
\usepackage{url}
\usepackage{tikz}
\usetikzlibrary{positioning,shapes}
\graphicspath{{./}{Fig/}}

\usepackage{hyperref}

\algnewcommand\algorithmicswitch{\textbf{switch}}
\algnewcommand\algorithmiccase{\textbf{case}}
\algnewcommand\algorithmicassert{\texttt{assert}}
\algnewcommand\Assert[1]{\State \algorithmicassert(#1)}%

\algnewcommand{\Initialize}[1]{%
  \State \textbf{Initialize:}
  \Statex \hspace*{\algorithmicindent}\parbox[t]{.8\linewidth}{\raggedright #1}
}
\algnewcommand{\Inputs}[1]{%
  \State \textbf{Inputs:}
  \Statex \hspace*{\algorithmicindent}\parbox[t]{.8\linewidth}{\raggedright #1}
}

\algnewcommand{\Try}[1]{%
  \State \textbf{Try:}
  \Statex \hspace*{\algorithmicindent}\parbox[t]{.8\linewidth}{\raggedright #1}
}

\algdef{SE}[SWITCH]{Switch}{EndSwitch}[1]{\algorithmicswitch\ #1\ \algorithmicdo}{\algorithmicend\ \algorithmicswitch}%
\algdef{SE}[CASE]{Case}{EndCase}[1]{\algorithmiccase\ #1}{\algorithmicend\ \algorithmiccase}%
\algtext*{EndSwitch}%
\algtext*{EndCase}%

\title{\LARGE \bf
Reactive and human-in-the-loop planning and control of multi-robot systems under LTL specifications in dynamic environments*
}

\author{Pian Yu$^{1}$, Gianmarco Fedeli$^{2}$, and Dimos V. Dimarogonas$^{3}$
\thanks{*This work was partially supported by the Swedish Research Council (VR), the Knut and Alice
Wallenberg Foundation (KAW), the EIC Horizon Europe SymAware, the EU
project CANOPIES, the ERC COG LEAFHOUND (Grant agreement ID: 864720), and the ERC ADG FUN2MODEL (Grant agreement ID: 834115). }
\thanks{$^{1}$Pian Yu is with the Department of Computer Science, Oxford University,
        Oxford, United Kingdom
        {\tt\small pian.yu@cs.ox.ac.uk}}%
\thanks{$^{2}$Fedeli Gianmarco is with Bosch, Braga, Portugal
        {\tt\small gianmarco.fedeli1@gmail.com}}%
\thanks{$^{3}$Dimos Dimarogonas is with the Division of Decison and Control Systems, KTH, Stockholm, Sweden
        {\tt\small dimos@kth.se}}%
}

\begin{document}

\maketitle
\thispagestyle{empty}
\pagestyle{empty}

\begin{abstract}

This paper investigates the planning and control problems for multi-robot systems under linear temporal logic (LTL) specifications. In contrast to most of existing literature, which presumes a static and known environment, our study focuses on dynamic environments that can have unknown moving obstacles like humans walking through. Depending on whether local communication is allowed between robots, we consider two different online re-planning approaches. When local communication is allowed, we propose a local trajectory generation algorithm for each robot to resolve conflicts that are detected on-line. In the other case, i.e., no communication is allowed, we develop a model predictive controller to reactively avoid potential collisions. In both cases, task satisfaction is guaranteed whenever it is feasible. In addition, we consider the human-in-the-loop scenario where humans may additionally take control of one or multiple robots. We design a mixed initiative controller for each robot to prevent unsafe human behaviors while guarantee the LTL satisfaction. Using our previous developed ROS software package, several experiments are conducted
to demonstrate the effectiveness and the applicability of the proposed strategies.

\end{abstract}

\section{INTRODUCTION}

During the past decade, there is a surge of using temporal logic formulas, such as linear temporal logic (LTL) \cite{Baier2008}, to concisely specify desired behaviors of robotic systems \cite{fainekos2009temporal,4141034,4459804,wongpiromsarn2012receding,ulusoy2013optimality,sahin2019multirobot,kantaros2020stylus,alonso2018reactive,guo2015multi}. This is due to the expressiveness of LTL formulas in capturing many common robotic tasks, e.g., ordered reachability, collision avoidance, surveillance, and ordered supply delivery \cite{guo2018probabilistic}. In addition, the availability of automated toolboxes \cite{Baier2008,belta2017formal} make the use of LTL more appealing. The planning and control under LTL specifications have been extensively investigated for a single robot \cite{fainekos2009temporal,4141034,4459804,wongpiromsarn2012receding} and multi-robot systems (MRSs) \cite{ulusoy2013optimality,sahin2019multirobot,kantaros2020stylus,alonso2018reactive,guo2015multi}. However, most of the existing literature assume that the environment is static and known a priori. When the environment is dynamic (e.g., there are moving obstacles like humans walking through), for which online replanning becomes a necessity, the problem of safe and efficient replanning under LTL specifications becomes challenging during the online execution of robots.

Along the rapid advancement of automation technology, the past decade has witnessed a growing emphasis on human-robot collaboration. Indeed, many autonomous systems are performing their designated tasks while being supervised or collaborating with human operators \cite{fong2003survey}. On one hand, having a human-in-the-loop is useful for guiding the robot
through challenging tasks. On the other hand, the possibly
erroneous inputs from the human
can be dangerous for the autonomous systems. Therefore, effectively managing real-time interactions between the autonomous system and humans is crucial for ensuring the safety and efficiency of the entire system. 

When a single robot is considered, authors in \cite{guo2013revising}
propose a plan revising mechanism assuming the environment
is static and partially known. Once a transition in the
current plan becomes invalid, it finds the shortest
path bridging up the two components.
The work \cite{lahijanian2016iterative} develops an iterative repair strategy to resolve unknown obstacles by combining local patching with
refined triangulation. However, this strategy cannot deal with
moving obstacles. For the case of MRSs, most of research assumes that the MRS adheres to a global LTL specification, subsequently addressing an offline motion planning problem through a centralized approach \cite{ulusoy2013optimality,sahin2019multirobot}. In \cite{yu2021distributed}, MRSs under local LTL specifications are
studied and a distributed 
motion coordination algorithm is proposed to resolve conflicts. In this work, the environment is static and local communication between robots is required for conflict resolution.  Furthermore, when humans are involved in the robot control, the concept of a mixed-initiative controller (MIC) is introduced in \cite{loizou2007mixed}, which integrates external human inputs with the conventional navigation controller. This method is further investigated in \cite{guo2018human}, where the human initiative influences both the high level tasks and the low level continuous inputs. 


This work aims to develop reactive planning and control algorithms for MRSs under LTL specifications. First, we build upon our prior research \cite{yu2021distributed} by addressing dynamic environments, which present a non-trivial challenge. Additionally, when humans participate in controlling the robots, we expand the MIC proposed in our previous work \cite{guo2018human} to accommodate MRSs. The contributions are threefold: i) Depending on
whether local communication is allowed between robots, we
propose two different online re-planning approaches for MRSs operating in dynamic environments. In the first scenario, where local communication among robots is allowed, we propose a local trajectory generation algorithm to resolve conflicts. In the second scenario, where there is no communication between the robots, we develop a model predictive controller to reactively handle potential collisions. In
both cases task satisfaction is guaranteed whenever it is feasible. ii) When
one or multiple robots are controlled by human operators, a MIC is designed for the MRSs to guarantee safety and LTL task satisfaction. iii) Several experiments are carried out to demonstrate the effectiveness
of the proposed strategies.


\section{Preliminaries and Problem Formulation}

\subsection{Linear temporal logic (LTL)}

We use LTL to concisely specify the desired robot behaviour. LTL is built from a set of atomic propositions $AP$, the logic connectives of negation ($\neg$), conjunction ($\wedge$), and disjunction ($\wedge$), and the temporal operators next ($\bigcirc$), until ($\mathsf{U}$), eventually ($\lozenge$), and always ($\square$). An LTL formula is formed inductively according to the following syntax \cite{Baier2008}:
\begin{equation}\label{LTL}
  \varphi::=\top|a|\neg\varphi|\varphi_1\wedge\varphi_2|\bigcirc \varphi|\varphi_1 \mathsf{U}\varphi_2,
\end{equation}
where $a\in AP, \varphi, \varphi_1, \varphi_2$ are LTL formulas. The logic connective $\vee$ and temporal operators $\lozenge$ and $\square$ can be derived inductively. We omit the full LTL semantics due to 
space limitations and refer the reader to \cite{Baier2008} for details. The satisfaction of an LTL formula $\varphi$ over $AP$ can be captured through a nondeterministic B\"{u}chi automaton (NBA) \cite{buchi1990decision}, defined as a tuple $\mathsf{B} = (S, S_{0}, 2^{AP}, \delta, F)$, where
\begin{itemize}
  \item $S$ is a finite set of states,
  \item $S_{0}\subseteq S$ is the set of initial states,
  \item $2^{AP}$ is the input alphabet,
  \item $\delta: S\times 2^{AP} \to 2^{S}$ is the transition function, and
  \item $F\subseteq S$ is the set of accepting states.
\end{itemize}
An infinite \emph{run}  $\bm{s}=s_0s_1\ldots$ is called \emph{accepting} if $\texttt{Inf}(s)\cap F\neq \emptyset$, where $\texttt{Inf}(s)$ is the set of states that appear in $\bm{s}$ infinitely often. Several translation tools, e.g., LTL2BA \cite{gastin2001},  are available to obtain $\mathsf{B}$ given $\varphi$.

\subsection{Robot dynamics and motion abstraction}

Consider a set of $N$ robots navigating within a bounded workspace $\mathcal{W}$. The dynamics of robot $i$ is given  by:
\begin{equation}\label{kinematics_eqn}
\begin{aligned}
\dot{x}_i(t) &= v_i(t)\cos(\theta_i(t)),\\
\dot{y}_i(t) &= v_i(t)\sin(\theta_i(t)),\\
\dot{\theta}_i(t) &= w_i(t), 
\end{aligned}
\end{equation}
where $p_i=[x_i, y_i]^T$ is the Cartesian position, $\theta_i$ is the orientation, and $u_i=[v_i, w_i]^T$ is the input vector of robot $i$. The input of robot $i$ is constrained to the compact set $\mathcal{U}_i$, i.e., $u_i(t)\in  \mathcal{U}_i, \forall t\ge 0.$ 

Within the workspace $\mathcal{W}$, there is a set of properties (atomic propositions) $AP=\left\{a_1, a_{2}, \cdots, a_{M}\right\}$, e.g. ''this is a goal region", ``this is a charging station". Let $\Phi:=\{X_1, \ldots, X_{M}\}$ be a partition of $\mathcal{W}$
such that $\mathcal{W}=\cup_{l=1}^{M} X_l$. Define $L: \mathcal{W}\to 2^{AP}$ as a labelling function. Given a point $p\in \mathcal{W}$, define the function $Q: \mathcal{W}\to \Phi$ as $Q(p):=\{X_l\in\Phi: p\in X_l\}.$ It
maps a state $p$ into the region $X_l$ that contains it. We say $\Phi$ is an \emph{observation preserving} partition of $\mathcal{W}$ if it satisfies 
\begin{equation*}
L(p)=L(p'), \forall p, p': Q(p)=Q(p').
\end{equation*}
Given an observation preserving partition of the workspace, the dynamics of robot $i$ (\ref{kinematics_eqn}) can be abstracted as a controlled  transition system (CTS) $\mathcal{T}_i=(\Phi, \phi_{0, i}, \mathcal{U}_i, AP, \rightarrow_i, L)$, where $\phi_{0, i}\in \Phi$ is the initial state and
$\rightarrow_i: \Phi\times \mathcal{U}_i \to  \Phi$ is the transition relation.

\subsection{Objectives}

We assign an LTL specification $\varphi_i$ to each robot $i$, which is specified over the set of atomic propositions $AP$. 
Nevertheless, the workspace is dynamic, for instance, there may be moving obstacles like humans walking through. In addition, it is assumed that each robot has only limited sensing capabilities, that is, the robot can detect other robots or the unknown moving obstacles only if they are within its sensing region $\mathcal{B}(p_i, R_i)$, where $p_i$ is the position and $R_i$ is the sensing radius of robot $i$. The initial trajectory planning (which is detailed in the next section) cannot take into account these unpredictable situations, and thus online replanning is necessary for each robot to guarantee safe operation and task satisfaction.

Given the abstract model, i.e., the CTS $\mathcal{T}_i$, and the task specification $\varphi_i$ of each robot $i$, the first objective (O1) is to design a planning algorithm for each robot $i$ such that the LTL task $\varphi_i$ is satisfied and safety is always guaranteed despite of the dynamic environments. In addition, we further consider the human-in-the-loop context, where human operators can take over the control of the robots
from the on-board autonomous controller. The second objective (O2) is to design control algorithm for each robot $i$, which can react to (possibly dangerous)
human inputs while preserving  safety and task satisfaction.

\section{Online Replanning in dynamic environments}

Before implementation, an initial satisfying plan needs to be generated for each robot $i$. This procedure is implemented in the package LTL core \& planner \cite{baran2021ros}. It is based on constructing a product B\"{u}chi automaton (PBA) $\mathcal{P}_i$ between the CTS $\mathcal{T}_i$ and the NBA $\mathsf{B}_i$ (which is translated from $\varphi_i$). 
Using model-checking methods \cite{belta2017formal}, an accepting run can be obtained from the PBA $\mathcal{P}_i$ and projected back to the CTS $\mathcal{T}_i$ intersection $\mathsf{B}_i$. Accepting runs have a \emph{prefix-suffix} structure of this kind: $r_{\mathcal{P}_i}=p_{0}, p_{1} \cdots p_{k}\left(p_{k+1} \ldots \ldots p_{n} p_{k}\right)^{\omega}$.
The output word is composed of two separate parts: a finite prefix that is executed only once from the initial state $p_{0}$ to an accepting state $p_{k}$ and a suffix that is repeated infinitely from the accepting state $p_{k}$ to itself.
The accepting run has a corresponding action sequence that each robot must carry out in a prefix-suffix structure in order to fulfill $\varphi_i$. 

Note that the initially planned trajectory for each robot $i$  does not account for the motion of other robots or changing of the environment. Thus, online replanning is necessary during the implementation. 
Two distinct cases are considered for online replanning, one is based on local communication between robots and the other assumes no communication between robots. In the former case, each robot has access to information (i.e., broadcast data from other robots) within its sensing region while the latter case does not. 

\subsection{Local communication case}

In this section, we consider that each robot can identify conflicts within its sensing region using the information broadcast by other robots. Before proceeding, the definition of a conflict is needed.

Let $p_i(t)=[x_i(t), y_i(t)]^T$ be the position of robot $i$ at time $t$. Denote by $p_i([t, t+\Delta_i])$ the local trajectory of robot $i$, where $\Delta_i=\min_{t'\ge t}\{p_i(t')\notin \mathcal{B}_i(p_i(t), R_i)\}$. We say there is a \emph{conflict} between robot $i$ and $j$ at time $t$ if there exists a region $X_k\in \Phi$ such that $p_i([t, t+\Delta_i])\cap X_k \neq \emptyset$ and $p_j([t, t+\Delta_j])\cap X_k \neq \emptyset$. This means that the local trajectories of robots $i$ and $j$ pass through the same region $X_k$.

Using the local trajectory information $p_j([t, t+\Delta_j])$ broadcast by the neighboring robots $j\in \mathcal{N}_i(t):=\{j: \|p_i(t)-p_j(t)\|\le R_i\}$, each robot $i$ can detect conflicts. Once conflicts are detected,   online replanning is conducted to ensure that conflicts are avoided and the LTL task is satisfied. The online replanning procedure consists of a local and a global trajectory generation algorithms, which are built upon our previous work \cite{yu2021distributed}. 

The local trajectory generation algorithm is detailed in Algorithm \ref{local_trajectory_algorithm}. Whenever conflicts are detected among robots, a priority hierarchy is first created to coordinate the planning order. Subsequently, a sampling-based algorithm is employed to create a local collision-free trajectory.
 
\vspace{0.4cm}
\begin{algorithm}
   \begin{algorithmic}[1]
\Require $\xi_i$, $\mathcal{B}_i$, $\mathcal{P}_i, V_{\mathcal{P}_{i}}, \mathbb{O}, \mathbb{O}_{conf}$
\Ensure A local transition system $\mathcal{T}_{i}^{L}$ and a leaf node $\xi_i^f$.
\State Initialize $\mathcal{T}_{i}^{L}=(S_i^{L}, S_{i,0}^{L}, \mathcal{U}_i, AP, \rightarrow_{i}^L, L)$ and $\xi_i^f=\emptyset$, where $S_i^{L} =S_{i,0}^{L}= \xi_i$ and $\rightarrow_{i}^L =\emptyset$.
\For {$k=1,\ldots, N_i^{\max}$},
\State $\xi_s\leftarrow \textit{generateSample}(SA_i)$,
\State $\xi_n\leftarrow \textit{nearest}(S_i^{L}, \xi_s)$,
\State Solve the optimization program $\mathcal{P}(\xi_n, \xi_s, \tau_s)$, which returns $(\xi_r, u_i^*)$,
\State $B_i(\xi_r)\leftarrow \textit{trackBuchiStates}(\mathcal{B}_i)$,
\If {$B_i(\xi_r)\neq \emptyset \wedge V_{\mathcal{P}_{i}}(\xi_r, B_i(\xi_r))<\infty$},
\If{$\operatorname{proj}_2\left(\left[\xi_n, \xi_r\right]\right) isObstaclesFree(\mathbb{O}, \mathbb{O}_{conf}) \wedge safeMotion(\xi_r, R_{safe})$},
\State $S_i^{L}\leftarrow S_i^{L}\cup\{\xi_r\}; \rightarrow_{i}^{L}= \rightarrow_{i}^{L}\cup \{\xi_n \xrightarrow[]{u_i^*} \xi_r\}$,
\EndIf
\EndIf
\If {$\texttt{proj}_2(\xi_r)\notin \mathcal{B}(p_i, R_i)$,}
\State $k=N_i^{\max}+1$,
\State $\xi_i^f\leftarrow \xi_r$,
\EndIf
\EndFor
\end{algorithmic}
    \caption{localTrajectoryGeneration}
    \label{local_trajectory_algorithm}
\end{algorithm}

Algorithm \ref{local_trajectory_algorithm} starts with randomly sampling the expanded sensing region of robot $i$. Then by constructing a local transition system, a collision-free trajectory is synthesized that simultaneously guarantees the fulfillment of the LTL task $\varphi_i$. 
It takes as input the current state $\xi_i=(x_i, y_i, \theta_i)$ of robot $i$, the offline constructed NBA $\mathsf{B}_i$ and PBA $\mathcal{P}_i$, the offline computed potential function $V_{\mathcal{P}_i}$ (the definition and computation can be found in \cite{yu2021distributed}), the set of known static obstacles $\mathbb{O}$, and the set of conflict regions $\mathbb{O}_{conf}^t$ as input.
Firstly, a local transition system $\mathcal{T}_{i}^L$ is initialized (line 1). At each iteration, a new state $\xi_{s}$ is taken randomly from the sampling area $SA_i:=\left\{(p, \theta): p \in \mathcal{B}\left(p_i, R_i+\eta\right)\right\}$ (line 3), where $\eta > 0$ is an offline constant which ensures that one can sample outside of the sensing region $\mathcal{B}\left(p_i, R_i\right)$, where $p_i$ is the current position of robot $i$. 
Then through the function $\textit{nearest}\left(S_i^L, \xi_s\right)$ an RRT primitive is applied, which returns the nearest state to $\xi_{s}$ in $S_i^L$ (line 4). 
At this stage an optimization problem $\mathcal{P}\left(\xi_n, \xi_s, \tau_s\right)$ is solved, so as to find the closest reachable state from the new sample $\xi_{s}$:
\begin{equation*}
    \hspace{-1cm} \min _{u_i \in \mathbb{U}_i}\left\|\xi_r-\xi_s\right\| 
\end{equation*}
\begin{equation*}
    \begin{aligned}
    \text { subject to: } 
    &\xi_i(0)=\xi_n, \\
    &\xi_r=\xi_n+\int_0^{\tau_{s}} F_i\left(\xi_i(s), u_i\right) ds,\\
    &u_i\in \mathcal{U}_i,
    \end{aligned}    
\end{equation*}
where $\tau_{s}$ represents the sampling time, while $F_i\left(\xi_i(s), u_i\right)$ describes the robot's dynamics  (\ref{kinematics_eqn}) (line 5). 
Once $\xi_r$ is obtained, the corresponding subset of valid B\"{u}chi states $B_i\left(\xi_r\right)$ is computed using algorithm \textit{trackBuchiState} (given in \cite{vasile2013sampling}, Algorithm 1) (line 6). If both conditions $B_i\left(\xi_r\right) \neq \emptyset$ and $V_{\mathcal{P}_{i}}\left(\xi_r, B_i\left(\xi_r\right)\right)<\infty$ are satisfied (which ensures the existence of a path originating from $\xi_r$ that leads to a self-reachable accepting state of  $\mathcal{P}_i$), then a potential new state is considered.
Such state $\xi_r$ is added into $S_i^L$ and the corresponding transition relation $\xi_n \stackrel{u_i^*}{\rightarrow} \xi_r$ is added into $\rightarrow_{i}^L$, if the path connecting $\xi_{n}$ with $\xi_{r}$ is obstacles free and the motion is considered safe. To verify these requirements, two new algorithms are designed in this work. The collision free requirement is checked through the algorithm   $\textit{isObstaclesFree}$, which computes the distance between the line segment $\operatorname{proj}_2\left(\left[\xi_n, \xi_r\right]\right)$ and the set of obstacles $\mathbb{O}\cup \mathbb{O}_{conf}$. If the line segment $\operatorname{proj}_2\left(\left[\xi_n, \xi_r\right]\right)$ is collision-free, we further check the safety of the motion using algorithm $safeMotion(\xi_r, R_{safe})$ (considering the base footprint of the robot), where $R_{safe}$ is the safe distance that we specified a priori (lines 7-11). The algorithm halts when the local sampling tree extends beyond the sensing area, and the leaf node $\xi_i^f$ is subsequently returned (as determined by the corresponding state $\xi_r$) (lines 12-15).

The global trajectory generation is similar to the initial trajectory generation, which uses the leaf node $\xi_i^f$ (which is obtained by Algorithm \ref{local_trajectory_algorithm}) as input of the 
 LTL core \& planner, and replans the sequence of actions in order to accomplish the LTL task $\varphi_i$.

\subsection{Communication-free case}
\label{comunnicatio_free_online_replan}
A communication-free scenario is also considered in pursuit of extending the work to other real-world scenarios where reliable communications between robots may not be available. A reactive collision avoidance algorithm is designed in combination with the local planner.

Before proceeding, the concept of ``trap state" is needed. \emph{Trap states} are PBA states from which the B\"{u}chi acceptance condition
cannot be fulfilled, i.e., states that cannot reach accepting
states that appear infinitely often. For the PBA $\mathcal{P}_i$ of robot $i$, the set of all trap states is
denoted $\mathbb{G}_i$. An algorithm for computing trap states is implemented in \cite{baran2021ros}.

In this case, the online replanning is activated for robot $i$ whenever obstacles (can be other robots and unknown static/moving obstacles) are detected within its sensing region $\mathcal{B}_i(p_i, R_i)$. Let $\mathbb{O}_{obs}^t$ be the set of moving obstacles detected at time $t$. 
In order to avoid trap states $\mathbb{G}_i$ (thus guaranteeing task feasibility), the set of known static obstacles $\mathbb{O}$, and the set of moving obstacles $\mathbb{O}_{obs}^t$ (thus guaranteeing safety), we consider a local model predictive controller:
\begin{equation}
\label{mpc_free_communication_eqn}
    \begin{aligned}
        \hspace{-0.5cm} \min _{u_i} \quad 
        & \int_{t=0}^{T} (\xi_i(t)- X_{des})^{T}Q(\xi_i(t)- X_{des}) + u_i^{T}Ru_i \\
        &+(\xi(T) - X_{des})^{T}Q_{N}(\xi(T)-X_{des})\hspace{0.3cm} \\
        & \hspace{-1cm} +\int_{t=0}^{T}{w_{\mathbb{O}_i}\frac{1}{dist(\xi(t), \mathbb{O}_{obs}^t\cup \mathbb{O})} + w_{\mathbb{G}_i}\frac{1}{dist(\xi(t), \mathbb{G}_i)}} \\
        \quad \textrm{subject to:  } 
        & \xi_i(t) = \xi_i(0)+\int_0^{t} F_i\left(\xi_i(s), u_i\right) ds, \\
        & u_i\in \mathcal{U}_i,\\
        & \xi_i(T) \in X_{des} \\
        & \xi_i(0)=\xi_i,
    \end{aligned}
\end{equation}
where $T$ is the horizon and $\xi_i$ is the current state. The  first 2 terms of the cost function constitute a quadratic cost of both state and input with the matrices $Q, R, Q_N$ being positive definite and $X_{des}\in \Phi$ denotes the goal state that one can choose based on the current satisfying trajectory (recall that the satisfying trajectory is discrete and have a prefix-suffix structure). The last term of the cost function considers both the distance between robot $i$ and the obstacles $\mathbb{O}_{obs}^t\cup \mathbb{O}$ (which is updated at each time step) and the distance between robot $i$ and the trap states $\mathbb{G}_i$, where the distance function is defined as $dist(x, A):=\inf_{y\in A}\{||x-y\|\}$ and $w_{\mathbb{O}_i}, w_{\mathbb{G}_i}$ are the corresponding weight parameters. For instance if one prioritizes the satisfaction of the given LTL task rather than not collide with obstacles, the weight $w_{\mathbb{G}_i}$ can be set higher than $w_{\mathbb{O}_{i}}$. 

\section{Human-in-the-loop control}

In this section, we consider the human-in-the-loop context, where the robots are assigned to complete complex tasks specified by an LTL formula, while a human operator can take over the control of the robot from the on-board autonomous controller. The robot is required to respect human inputs, but at the same time react to undesired (possibly dangerous) human behaviors in order to preserve safety and task satisfaction. This is useful for guiding the robot through challenging assignments.

The proposed strategy is a MIC with the intention of extending existing work  \cite{loizou2007mixed, guo2018human} to MRSs. The MIC, inspired by \cite{loizou2007mixed, guo2018human}, for robot $i$ is given by:
\begin{equation}\label{hil}
    u_i(t) \triangleq    
    u_i^r(t)+\kappa\left(r, \mathbb{O}, \mathbb{G}_i\right) u^h(t),
\end{equation}
where $u_i^r(t)$ is a given autonomous controller, the
function $\kappa\left(\xi_i, \mathbb{O}, \mathbb{G}_i\right) \in [0, 1]$ is a smooth function to be designed, and $u^h(t)$ is the human input function, which
is uncontrollable and unknown by the robot.

The autonomous controller $u_i^r(t)$ can be a function that navigates the robot from one region $X_s$ of the current discrete plan to the next one $X_g$ while staying within the workspace $\mathcal{W}$ when there are no conflicts/obstacles are detected. Otherwise, the local trajectory generation algorithm or the model predictive controller developed in Section III is implemented to get $u_i^r(t)$.
In order to guarantee the task satisfaction for all human inputs, the function $\kappa\left(\xi_i, \mathbb{O}, \mathbb{G}_i\right)$ is designed as:
\begin{equation*}
\begin{aligned}
    \kappa\left(\xi_i, \mathbb{O}, \mathbb{G}_i\right) \triangleq &G_\mathrm{mix}\cdot\frac{\rho\left(d_o-d_s\right)}{\rho\left(d_o-d_s\right)+\rho\left(\varepsilon+d_s-d_t\right)} \\
    &+ 
    (1 - G_\mathrm{mix})\cdot\frac{\rho\left(d_t-d_s\right)}{\rho\left(d_t-d_s\right)+\rho\left(\varepsilon+d_s-d_t\right)}
    \end{aligned}
\end{equation*}
where $d_t \triangleq \min _{\pi \in \mathbb{G}_i}\|\xi_i-\pi\|$ is the minimum distance between robot $i$ and any region within $\mathbb{G}_i$; $d_o \triangleq \min _{\pi \in \mathbb{O}_i^t}\|\xi_i-\pi\|$ is the minimum distance between the robot and any obstacle within $\mathbb{O}$; $\rho(s) \triangleq$ $e^{-1 / s}$ for $s>0$ and $\rho(s) \triangleq 0$ for $s \leq 0$, and $d_s, \varepsilon>0$ are design parameters as the safety distance and a small buffer. Moreover $G_\mathrm{mix} \in [0, 1]$ represents a gain parameter, in order to manage the trade-off between two aspects: preventing trap states and obstacle avoidance.

\section{Experimental results}

In this section, we present experimental studies to validate our results, which is also the main contribution of this paper. The proposed strategies hold potential for application in precision agriculture, enabling farmworkers to collaborate effectively with robot teams in carrying out agronomic tasks, such as harvesting or pruning in table-grape vineyards.

\begin{figure}[ht]
    \centering
    \includegraphics[width=0.15\textwidth]{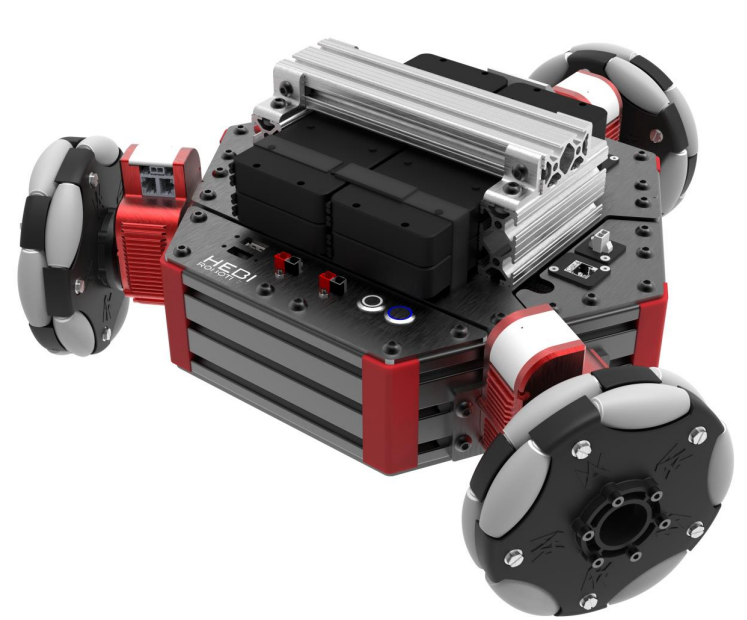}
    \includegraphics[width=0.2 \textwidth]{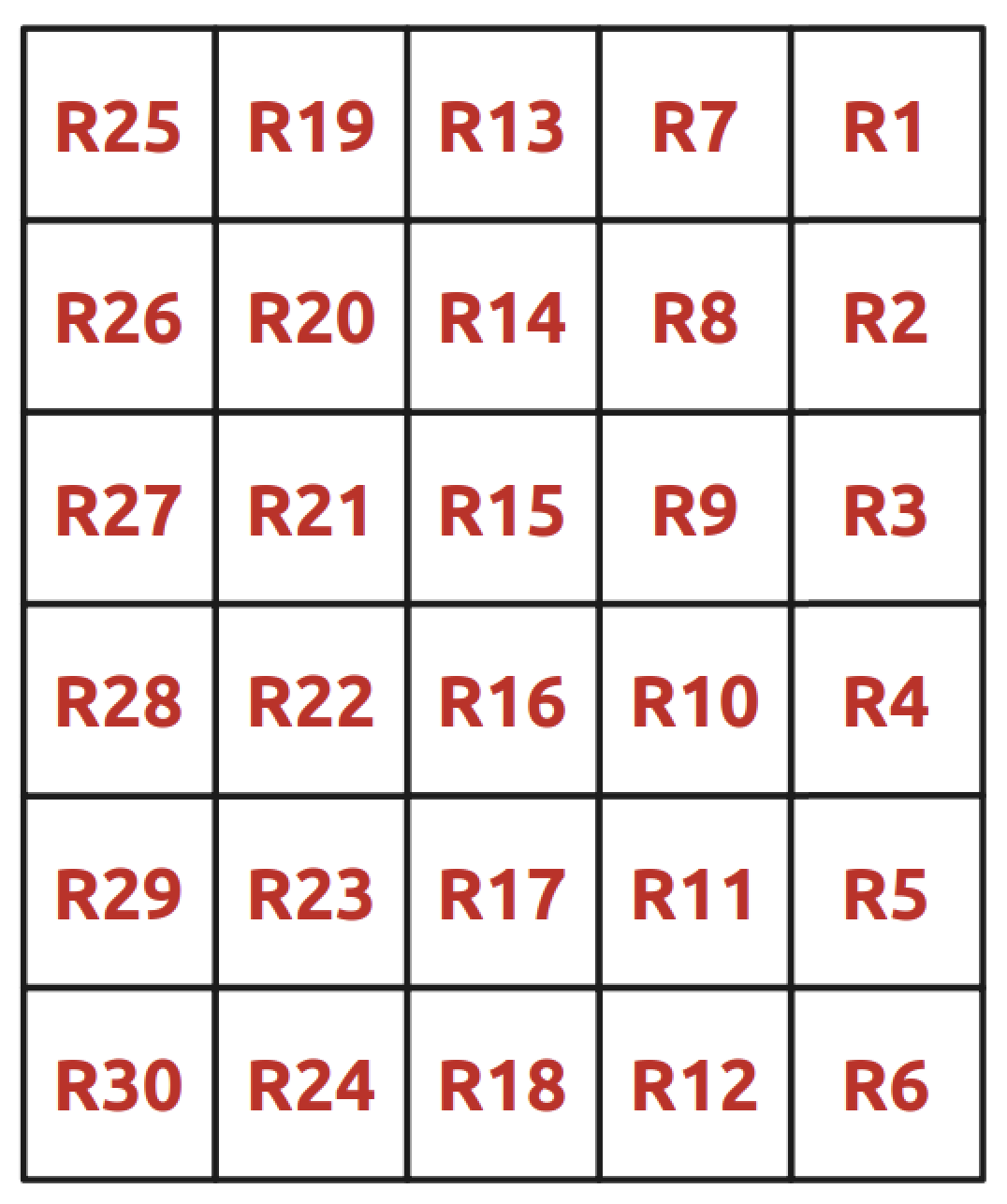}
    \caption{Rosie HEBI mobile base (left) and workspace discretization (right).}
    \label{fig:workspace_discretization}
\end{figure}

We use the Rosie HEBI mobile base (see Fig. \ref{fig:workspace_discretization} left), which is an omnidirectional mobile platform with three omnidirectional wheels. Through the use of the Qualisys motion capture system, the motion of the involved rigid bodies is tracked in real time. Each HEBI Rosie mobile base includes an on-board computer equipped with a proper ROS version, and autonomous control is accomplished from the user PC using the ROS API in conjunction with a wireless platform connection. 
The experiment workspace is an $5m \times 6m$ region as shown in Fig. \ref{fig:workspace_discretization} right, which is abstracted into 30 $1m\times 1m$ squares. We consider a group of 2 or 3 HEBI Rosie robots (Rosies 0, 1, and 2), and the dynamics of each robot is given by (\ref{kinematics_eqn}). In addition, each Rosie is subject to the inputs constraints $|v_i|\le 0.35 m/s$ and $|w_i|\le 0.35 rad/s$. The sensing radius of each robot is $R_i=0.8 m, \forall i$.

\subsection{Online replanning}

In this section, we consider $N=2$ HEBI Rosie robots. The LTL specification for each robot is given by
\begin{itemize}
    \item $\varphi_0=((\square \Diamond R8) \land (\square \Diamond R20))$,
    \item $\varphi_1=(\square \Diamond R17) \land (\square \Diamond R21)$.
\end{itemize}

\subsubsection{Local communication case}

\begin{figure}[ht!]\centering
{\label{a}\includegraphics[width=.8\linewidth]{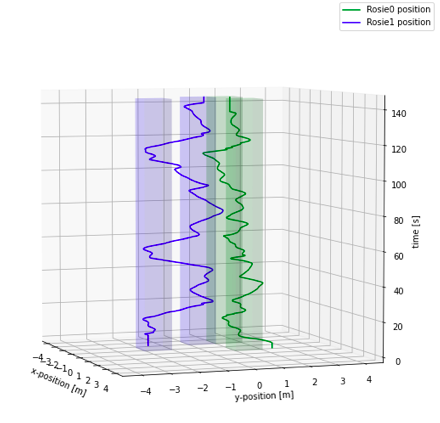}} 
\caption{Local communication case: the real-time position trajectories of Rosies 0 (green line) and 1 (purple line). The light green and purple regions represent the target regions of Rosies 0 and 1, respectively.}
\label{figure_exp2_local_communication0}
\end{figure}

Initially each robot synthesizes a satisfying trajectory using the LTL core \& planner \cite{baran2021ros}. 
During the online implementation, possible collisions are detected, and then the local trajectory generation algorithm (Algorithm 1) is executed taking into consideration the  future trajectories of neighboring robots (which is acquired through local communication). The real-time position trajectories of the 2 Rosies are depicted in Fig. \ref{figure_exp2_local_communication0}.
During the experiment time horizon $140 s$, both robots successfully complete the surveillance tasks. Conflicts are detected 4 times in total, and the local trajectory generation algorithm is activated to resolve them every time. Note also that the input constraints are satisfied at all time. A video demonstration of this experiment can be found at \url{https://www.youtube.com/watch?v=mKWpqvMrW9Y}. 

\subsubsection{Communication-free case with human as moving obstacle}
In this case, we additionally consider a human  walking in the workspace $\mathcal{W}$ to evaluate the effectiveness of the model predictive controller.


\begin{figure}[ht!]\centering
{\label{a}\includegraphics[width=.8\linewidth]{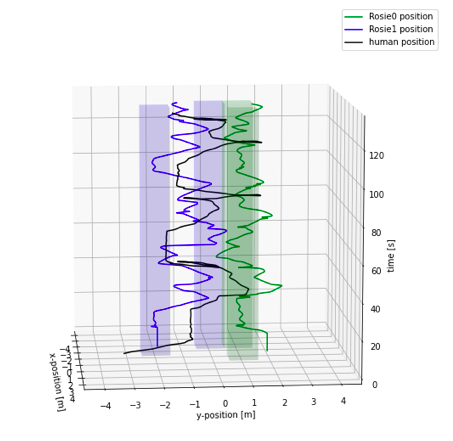}}
\caption{Communication-free case: the real-time position trajectories of Rosies 0 (green line), 1 (purple line), and human (black line). }
\label{figure_exp2_free_communication0}
\end{figure}


Fig. \ref{figure_exp2_free_communication0} depicts the real-time position trajectories of the 2 Rosies and the human, where the proposed model predictive controller (\ref{mpc_free_communication_eqn}) is applied for both Rosies. Differently from the local communication case where replanning is activated only if conflicts are detected, in the communication-free case, the online replanning is activated whenever a robot detects obstacles in its sensing region. As a result, the motion of each robot is affected not only by other robots, but also by the human randomly walking through the workspace.

Replanning is conducted 14 times in total over the experiment time horizon $120 s$, and one can see that the model predictive controller (\ref{mpc_free_communication_eqn}) guarantees the task satisfaction and collision avoidance (with other robots and the moving human) for each robot.
When compared to the local communication case, the number of replanning is higher since the replanning process is activated whenever obstacles are detected and the model predictive controller does not take into account future trajectories of other robots.
 A video demonstration of this experiment can be found at \url{https://youtu.be/TYgfbrk7hDs}.


\subsection{Human in-the-loop control}

\begin{figure}[ht]
    \centering
    \includegraphics[scale=0.18]{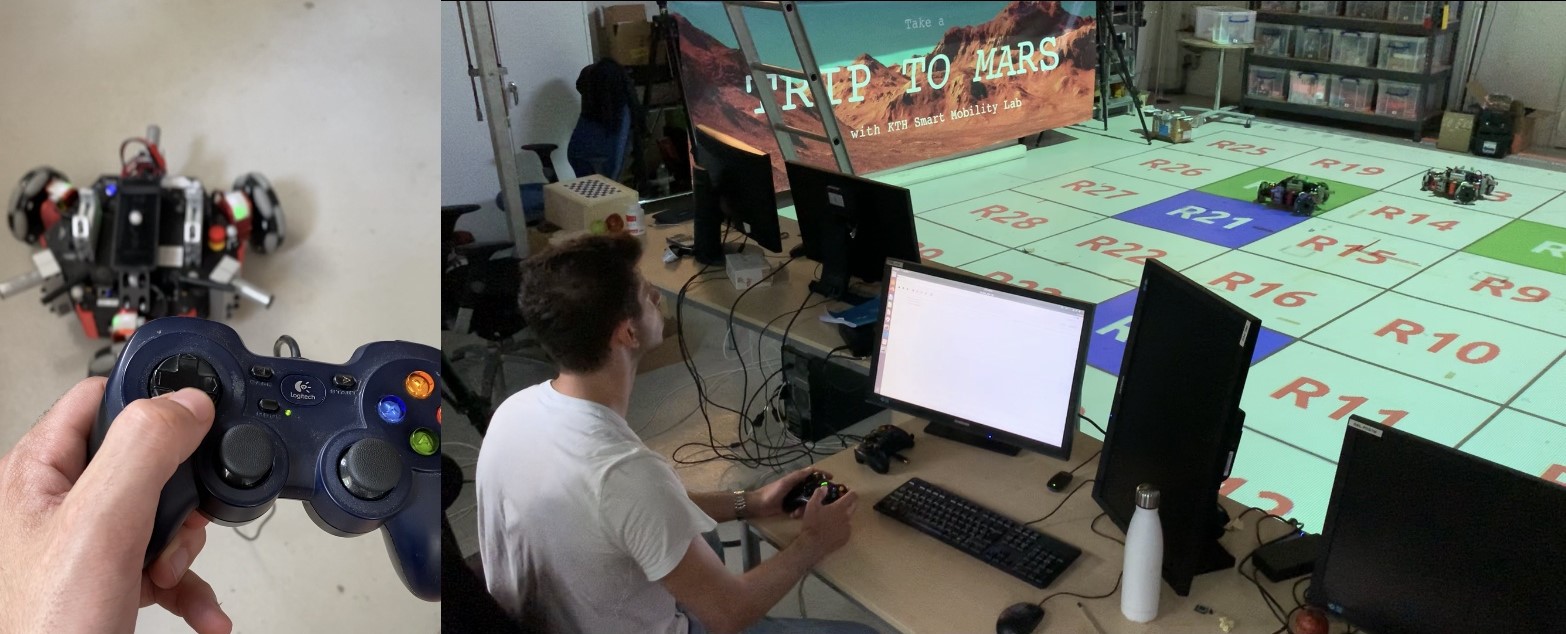}
    \caption{Human in-the-loop control.}
    \label{fig:hil_scenario}
\end{figure}

This section aims to evaluate the MIC explained in Section V in the human in-the-loop context, where a human may take control of one or more robots during the motion. We consider $N=3$ HEBI Rosie robots and the LTL specification for each robot is given by:
\begin{itemize}
    \item $\varphi_0=(\square \Diamond R8) \land (\square \Diamond R20)$,
    \item $\varphi_1=(\square \Diamond R22) \land (\square \Diamond R28)$, 
    \item $\varphi_2=(\square \Diamond R10) \land (\square \Diamond R11)$.
\end{itemize}

\begin{figure}[ht!]\centering
{\label{a}\includegraphics[width=.8\linewidth]{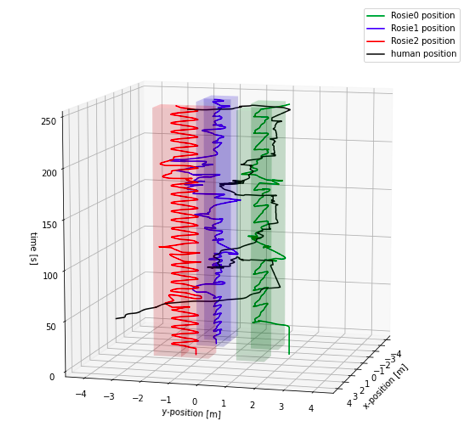}}
\caption{\label{figure_exp3_hil1_0} Human-in-the-loop: the real-time position trajectories of Rosies 0 (green line), 1 (purple line), 2 (red line), and human (black line), where Rosie 1 is controlled by the human operator. The light green, purple, and red regions represent the target regions of Rosies 0, 1, and 2, respectively.}
\end{figure}

Each robot can be controlled by a human with control inputs executed through a joystick equipped with bluetooth, see Fig. \ref{fig:hil_scenario}. The human operator, on one hand, influences robot autonomy when evaluating the performance of the proposed MIC, where the human inputs and autonomous inputs are fused, as detailed in Section V. On the other hand, the human operator is also walking through the workspace as a moving obstacle.  For the first $60s$, all robots are not affected by the human obstacle and no joystick inputs are applied. After that, the human starts to control Rosie 1 and walk through the workspace, and the model predictive controller (communication-free case) is applied to deal with potential conflicts for each robot. Additionally, some dangerous human behaviours are tested, including the human attempting to guide the robot towards itself. The real-time position trajectories of the 3 Rosies and the human are depicted in Fig. \ref{figure_exp3_hil1_0}. One can see that the MIC (\ref{hil}) prevents undesired actions and ensures safety.  A video demonstration of this experiment can be found at \url{https://youtu.be/2hWe5Wu52Bg}.

\section{Conclusions}

The work focuses on the design of efficient planning and control algorithms for MRSs that are subject to LTL specifications. Due to the
practical assumption of local sensing and dynamic environments, online replanning is needed for each robot to guarantee safety and task satisfaction. Two distinct cases are considered for online replanning. The former one assumes local communication between robots and a local trajectory generation algorithm is proposed to resolve conflicts.  The latter one assumes no communication between robots and a model predictive controller is designed to deal with penitential collisions. 
Finally, the human-in-the-loop context is considered
 a MIC is adopted to prevent unsafe human behavior. 

\addtolength{\textheight}{-12cm}   






\end{document}